\pdfoutput=1

\documentclass[11pt]{article}

\usepackage[]{acl}

\usepackage{times}
\usepackage{latexsym}

\usepackage[T1]{fontenc}

\usepackage{graphicx}
\usepackage{booktabs}
\usepackage{multirow}
\usepackage{rotating}
\usepackage{amsmath}
\usepackage{bm}
\usepackage{framed}
\usepackage{url}
\usepackage{caption}
\usepackage{xcolor}
\usepackage{algorithm}
\usepackage[noend]{algorithmic}
\usepackage{pifont}
\usepackage{soul}

\newcommand{\tabincell}[2]{\begin{tabular}{@{}#1@{}}#2\end{tabular}}
\definecolor{msgrblue}{HTML}{4889f4}
\definecolor{msgrgray}{HTML}{e1e1e7}
\newcommand{\win}[1]{{\color{white}{\sethlcolor{msgrblue}\hl{\textbf{#1}}}}}
\newcommand{\lose}[1]{\sethlcolor{msgrgray}\hl{#1}}

\definecolor{green}{HTML}{009933}
\definecolor{orange}{HTML}{ff9900}
\definecolor{lightpink}{HTML}{ffb6c1}
\definecolor{lightblue}{HTML}{add8e6}
\definecolor{lightsalmon}{HTML}{ffa07a}
\definecolor{lightgreen}{HTML}{90ee90}

\usepackage[utf8]{inputenc}

\usepackage{microtype}

\title{Exploring Prompt-based Few-shot Learning for \\ Grounded Dialog Generation}

\author{Chujie Zheng, Minlie Huang\\
  The CoAI Group, DCST, Institute for Artificial Intelligence, \\
  State Key Lab of Intelligent Technology and Systems, \\
  Beijing National Research Center for Information Science and Technology, \\
  Tsinghua University, Beijing 100084, China \\
  {\tt chujiezhengchn@gmail.com, aihuang@tsinghua.edu.cn} \\
}

\begin{document}
\maketitle
\begin{abstract}
Dialog models can be greatly strengthened through grounding on various external information, but grounded dialog corpora are usually not naturally accessible.
In this work, we focus on the few-shot learning for grounded dialog generation (GDG).
We first propose a simple prompting method for GDG tasks, where different constructs of model input, such as the grounding source and the conversation context, are distinguished through continuous or discrete prompts.
On three typical GDG tasks, we empirically demonstrate and analyze in-depth the effectiveness of our method.
We then conduct extensive experiments to thoroughly investigate how our prompting method works with different pre-trained models.
We show that prompted language models perform superiorly to conversational models, and further analyze various factors that influence the effects of prompting.
Overall, our work introduces a prompt-based perspective to the few-shot learning for GDG tasks, and provides valuable findings and insights for future research.
\end{abstract}

\section{Introduction}

Previous works have greatly enhanced dialog models through grounding model-generated dialogs on various external information \cite{ghazvininejad2018knowledge, huang2020challenges}, such as Wikipedia documents \cite{dinan2018wizard}, persona descriptions \cite{zhang-etal-2018-personalizing} or emotional support strategies \cite{liu-etal-2021-towards}.
However, grounded dialog corpora usually do not naturally exist and are mostly collected via crowd-sourcing, which could restrict the scale of accessible data.
Hence, the ability of \textit{few-shot learning}\footnote{
In the few-shot learning setting of this work, we assume that \textit{only a small amount of data samples are accessible and no additional data is used}, which is thus distinct from works that address the low-resource learning via pre-training on extra corpora \cite{Zhao2020Low-Resource, li-etal-2020-zero, liu-etal-2021-three}.
}
becomes increasingly necessary for grounded dialog models.

Compared to general dialog generation, where the response is only conditioned on the conversation context, grounded dialog generation (GDG) contains the other condition: the grounding source (GS).
We regard that this additional condition brings two major challenges to GDG tasks.
First, the models need to discriminate the more complex input constructs (not only utterances from different speakers, but also distinct input components, i.e., the GS and the conversation context).
Second, the concept of ``grounding'' is too abstract for models to grasp the relationship between the target response and the GS and further learn how to use the information of the GS.
These challenges are even more intractable under the few-shot setting.

\begin{figure}[t]
  \centering
  \includegraphics[width=\linewidth]{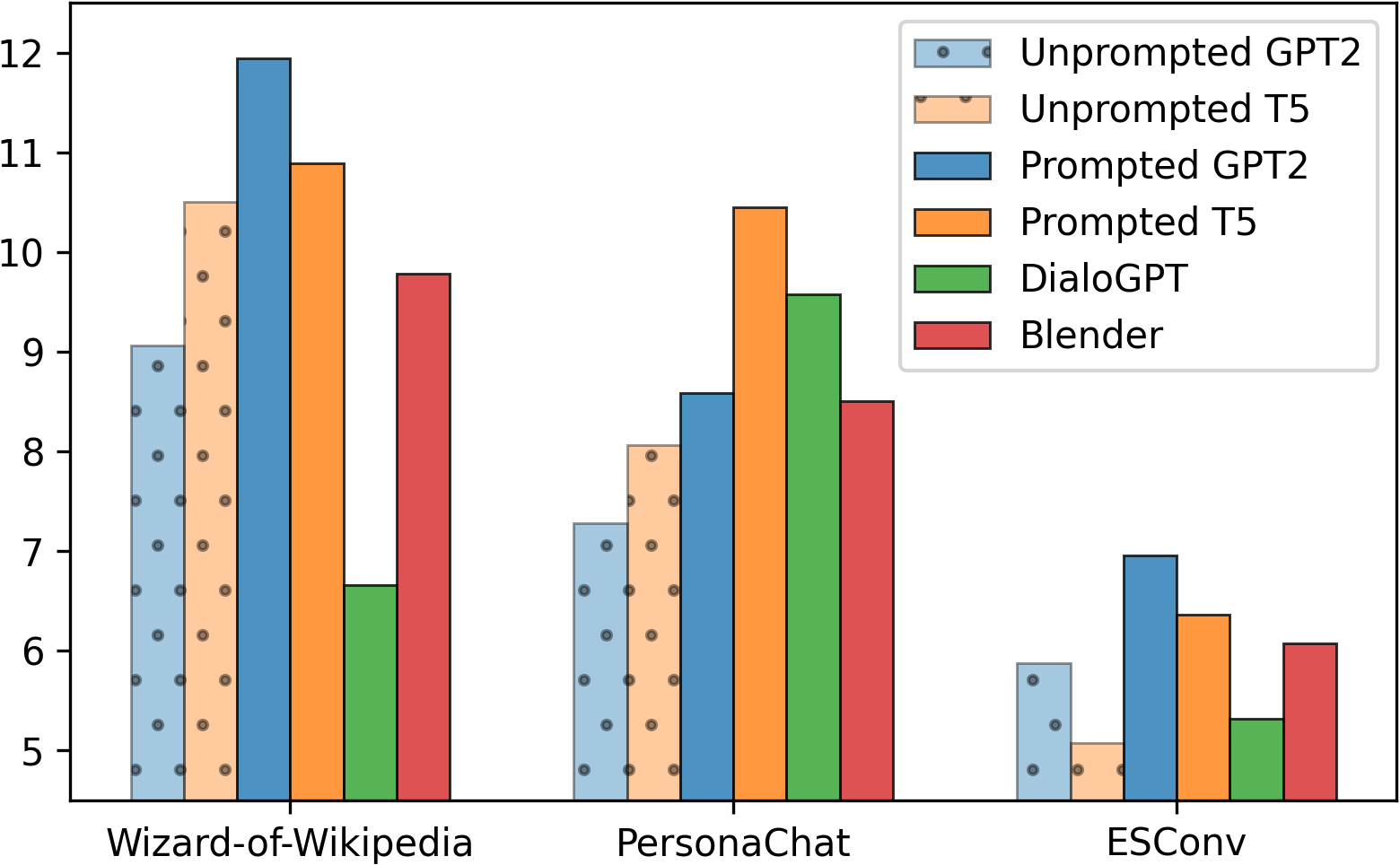}
  \caption{
  Overview of the effects of prompting.
  Model performance is measured by BLEU-2.
  Full experimental results are shown in Table \ref{tab:comparison}.
  Detailed experimental setups are described in \S \ref{sec:setup}.
  }
  \label{fig:intro}
  \vspace{-2mm}
\end{figure}

\begin{figure*}[t]
  \centering
  \includegraphics[width=\linewidth]{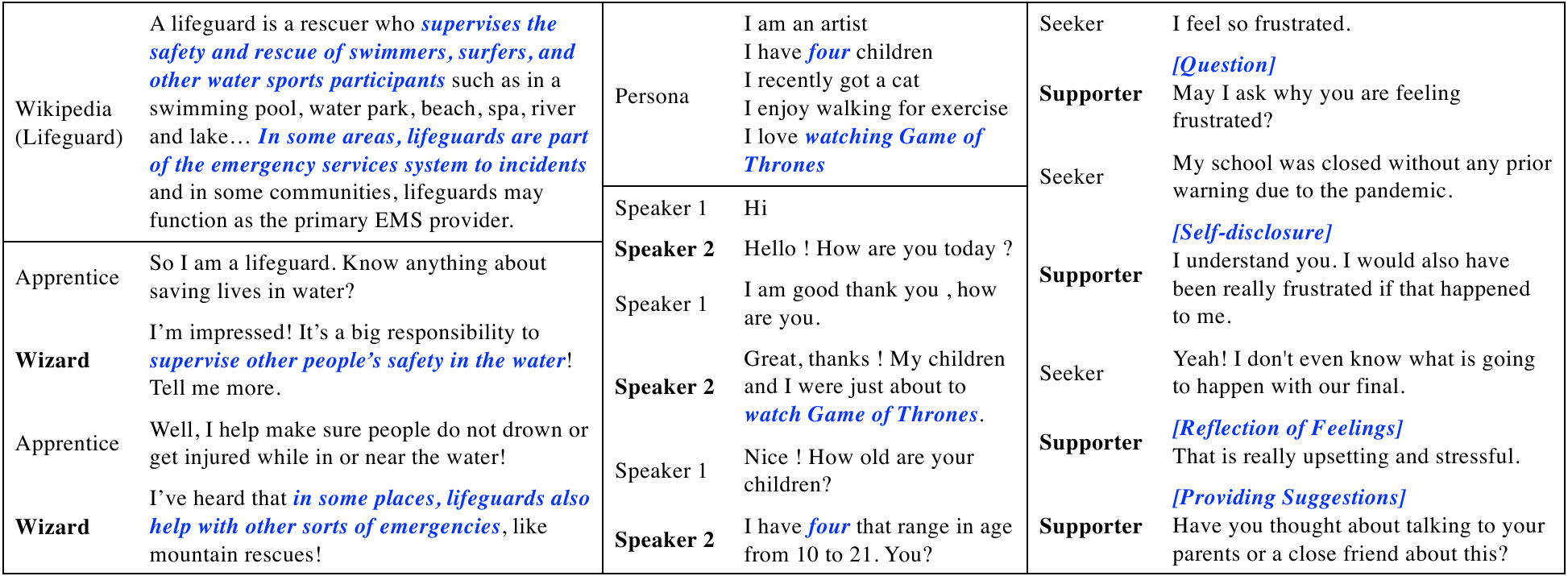}
  \caption{
  Examples of grounded dialogs from Wizard-of-Wikipedia \cite{dinan2018wizard}, PersonaChat \cite{zhang-etal-2018-personalizing} and ESConv \cite{liu-etal-2021-towards}, which are grounded on the Wikipedia knowledge (\textbf{left}), the persona profile (\textbf{middle}) and the emotional support strategies (\textbf{right}) respectively.
  The parts of grounding sources that the utterances are engaged in are marked in \textbf{\textit{\textcolor{blue}{blue}}}.
  }
  \label{fig:data_example}
  \vspace{-2mm}
\end{figure*}

Inspired by recent advances in pre-trained models and prompt-based learning \cite{liu2021pre}, which has shown impressive results in few-shot learning for various NLP tasks, in this paper we in depth explores prompt-based few-shot learning for grounded dialog generation.
As far as we know, this work is the first attempt that applies the prompting method to boost the few-shot learning performance for GDG tasks.
Our contributions fall into the following two aspects.

\begin{figure*}[t]
  \centering
  \includegraphics[width=0.9\linewidth]{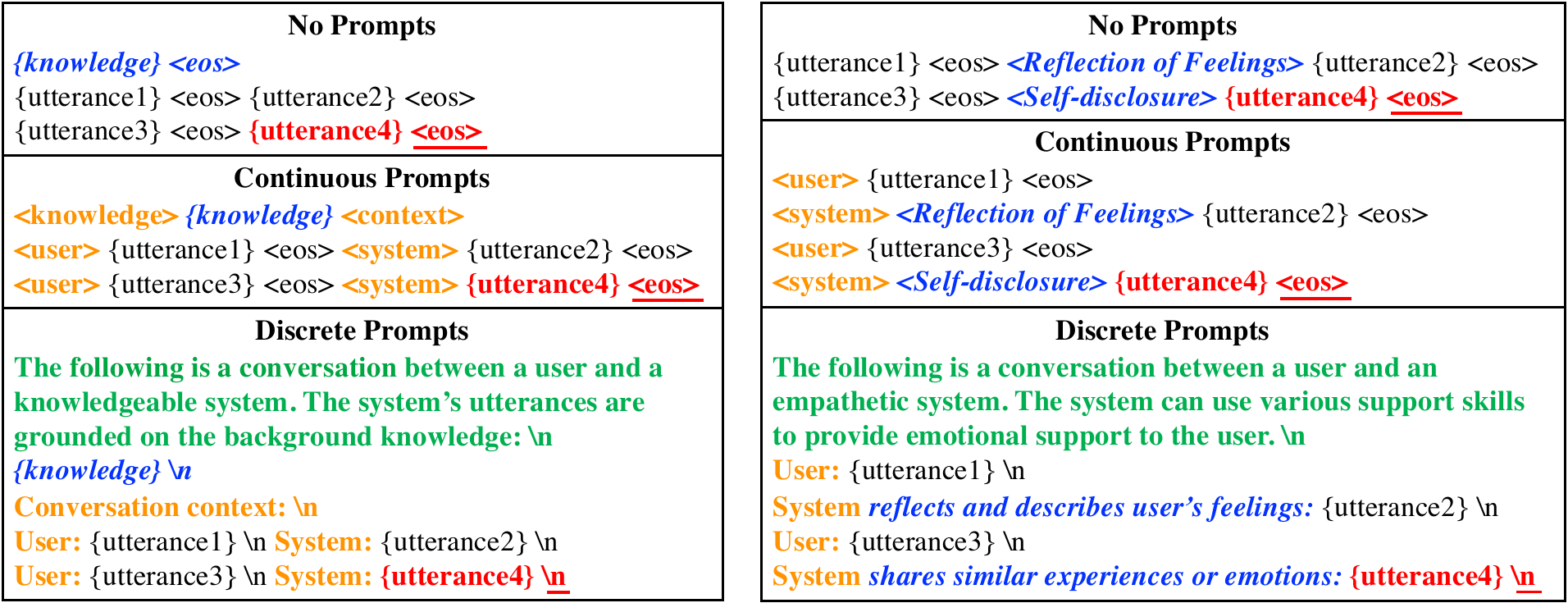}
  \caption{
  Illustration of our prompting method.
  Our method inserts \textbf{\textcolor{orange}{indicative tokens}} (\textbf{middle, Continuous Prompts}) or \textbf{\textcolor{orange}{textual descriptions}} (\textbf{lower, Discrete Prompts}) into input sequences to distinguish different input constructs.
  For the discrete prompts, we also prepend input sequences with \textbf{\textcolor{green}{task instructions}}.
  All the model parameters are trained with the negative log likelihood loss on the \textbf{\textcolor{red}{target responses}} (the end token, either ``<eos>'' or the line break ``\textbackslash n'', is \textbf{\textcolor{red}{\underline{underlined}}}).
  }
  \label{fig:method}
  \vspace{-2mm}
\end{figure*}

\textbf{First}, we propose a simple prompting method for GDG tasks, where the complex input constructs (i.e., distinct input components and different speakers' utterances) are distinguished through \textit{continuous} or \textit{discrete prompts}, as illustrated in Figure \ref{fig:method}.
Taking GPT2-medium as the backbone model, we empirically verify and analyze the effectiveness of our proposed method (\S \ref{sec:gpt2}).

\textbf{Second}, we conduct extensive experiments to thoroughly investigate how our prompting method works with different pre-trained models.
Results demonstrate that prompted language models (e.g., GPT2 and T5) can achieve superior performance to conversational models (e.g., DialoGPT and Blender) (Figure \ref{fig:intro} and \S \ref{subsec:vs}), and various factors also influence the effects of prompting (\S \ref{subsec:analysis}).
The key findings in our work are summarized in \S \ref{sec:findings}.

\section{Related Work}
\label{sec:related}

\noindent \textbf{Pre-training and Prompt-based Learning}
\quad
Recently, pre-trained models have shown the dramatic utility in various NLP tasks \cite{devlin-etal-2019-bert, radford2019language, raffel2020exploring}, which learn general-purpose language representation through pre-training on massive textual data with unsupervised learning objectives.
The prompt-based learning further takes the power of pre-trained models to unprecedented heights, especially in terms of few-shot learning \cite{brown2020language}.
In this paradigm, the pre-trained models are stimulated to solve downstream tasks through inserting discrete or continuous prompts into either original model inputs \cite{schick-schutze-2021-shot} or hidden states \cite{li-liang-2021-prefix}.
We refer readers to \cite{liu2021pre} for a comprehensive survey.

\noindent \textbf{Grounded Dialog Generation (GDG)}
\quad
In the past few years, researchers are increasingly interested in grounding machine-generated dialogs on various external information \cite{ghazvininejad2018knowledge, zhou-2018-commensense, zhou-etal-2018-dataset, Gopalakrishnan2019, zheng-etal-2020-difference, zhou-etal-2020-kdconv}.
As shown in Figure \ref{fig:data_example}, \cite{dinan2018wizard} utilizes Wikipedia documents as the background knowledge.
\cite{zhang-etal-2018-personalizing} equips conversational agents with pre-defined persona profiles to make them more engaging.
\cite{liu-etal-2021-towards} grounds on diverse emotional support strategies, enabling dialog models to be more empathetic and to provide more effective emotional support.

\noindent \textbf{Low-resource Learning for GDG}
\quad
Leveraging pre-training techniques, recent works also made attemps to address GDG tasks under a low-resource setting \cite{Zhao2020Low-Resource, li-etal-2020-zero, liu-etal-2021-three}.
Our work is distinguished from these in that instead of facilitating downstream fine-tuning via pre-training on extra corpora, we focus on making the most use of accessible data samples to perform few-shot learning.
While one can expect that combining our prompting method with previously adopted pre-training techniques would lead to better few-shot learning performance, we do not specially evaluate this but leave it for future work.

\section{Key Findings}
\label{sec:findings}

Our work evaluates the proposed prompting method (\S \ref{sec:gpt2}) and investigates its effectiveness with different pre-trained models (\S \ref{sec:comparison}).
The key findings are summarized as follows.

\noindent 
\textbf{1. Distinguishing the input constructs is an effective approach to boost the few-shot learning performance for GDG tasks (\S \ref{subsec:continuous}).}
However, poorly initialized prompts would instead damage model learning and lead to poor performance.

\noindent 
\textbf{2. Discrete prompts generally outperform continuous prompts under both few-shot and full-data settings (\S \ref{subsec:discrete}).}
In particular, minor perturbations do not result in significant performance fluctuation.
It indicates the practicability of manually crafted discrete prompts and that it may be not necessary to optimize them laboriously.

\noindent
\textbf{3. Prompted language models are superior to conversational models (\S \ref{subsec:vs}).}
Despite that our prompting method greatly benefits language models, it does not work with conversational models.

\noindent
\textbf{4. Our prompting method works across different model architectures, while its effectiveness also relies on backbone models with enough prowess (\S \ref{subsec:analysis}).}
Specifically, prompting is especially effective if the backbone models have large enough sizes and are pre-trained with general pre-training objectives (e.g., language modeling).

\section{Experimental Setups}
\label{sec:setup}

\subsection{Data Preparation}
\label{subsec:data}

Our experiments were conducted on three typical GDG tasks: Wizard-of-Wikipedia, PersonaChat and ESConv.
Their data examples are shown in Figure \ref{fig:data_example}, and the statistics are listed in Table \ref{tab:stats}.

\begin{table}[t]
  \centering
  \scalebox{0.75}{
    \begin{tabular}{cc|c|c|c}
    \toprule
          &       & \textbf{WoW} & \textbf{PC} & \textbf{ESConv} \\
    \midrule
    \multicolumn{5}{c}{\textit{\textbf{Data Split}}} \\
    \midrule
    \multirow{2}[0]{*}{\textit{Full-data}} & Train & 66K   & 50K   & 12K \\
          & Valid & 7K    & 6K    & 3K \\
    \midrule
    \multirow{2}[0]{*}{\textit{Few-shot}} & Train & \multicolumn{3}{c}{50} \\
          & Valid & \multicolumn{3}{c}{15} \\
    \midrule
    \multicolumn{2}{c|}{Test} & \multicolumn{3}{c}{3K} \\
    \midrule
    \multicolumn{5}{c}{\textit{\textbf{Sequence Length}}} \\
    \midrule
    \multicolumn{2}{c|}{\tabincell{c}{Context \\ (5 Utterances)}} & 68.5$_{\le \text{269}}$ & 52.3$_{\le \text{95}}$ & 90.2$_{\le \text{427}}$ \\
    \multicolumn{2}{c|}{Grounding Source} & 224.0$_{\le \text{368}}$ & 34.5$_{\le \text{53}}$ & - \\
    \multicolumn{2}{c|}{Response} & 22.6$_{\le \text{90}}$ & 13.3$_{\le \text{25}}$ & 21.6$_{\le \text{155}}$ \\
    \midrule
    \multicolumn{5}{c}{\textit{\textbf{Truncation Length}}} \\
    \midrule
    \multicolumn{2}{c|}{Context} & 250   & 150   & 250 \\
    \multicolumn{2}{c|}{Grounding Source} & 300   & 100   & - \\
    \multicolumn{2}{c|}{Response} & 50    & 25    & 50 \\
    \bottomrule
    \end{tabular}%
    }
  \caption{
  Data statistics.
  Each (context, response) pair is viewed as one data sample.
  Sequence lengths are calculated on the test sets (with GPT2 tokenization).
  Subscripts ``$\le x$'' denotes the max length in the test set is $x$.
  The contexts are truncated at the left while the grounding sources and the responses are at the right.
  }
  \label{tab:stats}%
  \vspace{-2mm}
\end{table}%

\noindent \textbf{Wizard-of-Wikipedia (WoW)} \cite{dinan2018wizard} 
is a knowledge-grounded dialog task, where the model makes use of Wikipedia documents to converse and provide information.
In WoW, each model-side utterance either refers to a knowledge sentence from the first paragraph of the selected Wikipedia entry or does not refer to any knowledge.
We removed the data samples where the responses do not use knowledge reference, and used the first paragraph of the selected Wikipedia entry as the GS for each sample.

\noindent \textbf{PersonaChat (PC)} \cite{zhang-etal-2018-personalizing}
is a persona-grounded dialog task, where the model is assigned with a pre-defined profile consisting of several textual persona descriptions.
We removed the data samples where the responses do not have any non-stop word overlap with the persona profiles (using the NLTK stop word list).

\noindent \textbf{ESConv} \cite{liu-etal-2021-towards}
is a support strategy-grounded dialog task, where the model uses various support strategies to provide emotional support to the help-seekers.
Note that different from WoW and PC where GS is in the form of unstructured texts, ESConv takes discrete concepts (support strategies) as the GS, which are more abstract and have more complex meanings.

WoW and PC adopted the official data split, while ESConv was manually split into 12K/3K/3K.
Note that for the sake of experimental efficiency, for WoW and PC we held 3K test samples\footnote{
Due to that WoW has an in-domain and the other out-of-domain test set, we held 1.5K samples from each set to construct the whole test samples (totally 3K).
}.
For the few-shot setting, we randomly sampled 50/15 data samples from the original training/validation sets, using the proportions in \cite{li-liang-2021-prefix}.
We did four random samplings to obtain four different subsets, and ran two random seeds for each subset.
Consequently, \textit{each reported final experimental result was obtained by averaging on eight (4*2) different original results}.

\subsection{Implementation Details}
\label{subsec:implementation}

\noindent \textbf{Training} \quad 
We trained all the model parameters during fine-tuning.
Unless otherwise specified, for the few-shot setting and for all the models and all the tasks, we employed the AdamW \cite{loshchilov2018decoupled} optimizer with batch size 5 and learning rate 2e-5, and used the linear learning rate scheduler with warmup steps 5. 
Gradient checkpointing was applied to reduce GPU memory occupation.
Models were trained for 10 epochs, and checkpoints were selected based on the perplexity on vaidation sets.
For the full-data setting, the learning rate and training epoch number were 1e-5 and 5 respectively.

\noindent \textbf{Inference} \quad
For WoW, following \cite{Zhao2020Low-Resource, li-etal-2020-zero}, we employed beam search with a beam size 3.
For PC and ESConv, following \cite{wolf2019transfertransfo,  liu-etal-2021-towards}, we additionally adopted Top-$p$ sampling \cite{holtzman2019curious} (temperature $\tau=0.7$ and $p=0.9$).
For WoW and ESConv, the min/max generation lengths were 10/50 respectively, while PC was 5/25.

\begin{table*}[t]
  \centering
  \scalebox{0.75}{
    \begin{tabular}{l|cccc|cccc|cccc}
    \toprule
       & \multicolumn{4}{c|}{\textit{\textbf{Wizard-of-Wikipedia}}} & \multicolumn{4}{c|}{\textit{\textbf{PersonaChat}}} & \multicolumn{4}{c}{\textit{\textbf{ESConv}}} \\
\cmidrule{2-13}       & \textbf{PPL ↓} & \textbf{B-2} & \textbf{F1} & \textbf{Wiki F1} & \textbf{PPL ↓} & \textbf{B-2} & \textbf{F1} & \textbf{PSN F1} & \textbf{PPL ↓} & \textbf{B-2} & \textbf{F1} & \textbf{Match} \\
    \midrule
    \midrule
       & \multicolumn{4}{c|}{\textit{\textbf{Full-data (66K)}}} & \multicolumn{4}{c|}{\textit{\textbf{Full-data (50K)}}} & \multicolumn{4}{c}{\textit{\textbf{Full-data (12K)}}} \\
    \midrule
    w/o GS & 17.7 & 8.5$^{**}$ & 21.9$^{**}$ & 4.0$^{**}$ & 15.2 & 10.9$^{**}$ & 25.3$^{**}$ & 6.6$^{**}$ & 15.1 & 6.6$^{**}$ & 20.6$^{**}$ & 20.7$^{**}$ \\
    Random & 9.0  & \textbf{15.5} & \textbf{28.0} & \textbf{9.1} & 11.4  & \textbf{12.2} & 26.6  & \textbf{11.8} & 14.7  & 7.6$^{*}$ & 22.5  & 40.6$^{**}$ \\
    Semantic & \textbf{9.0} & 15.3  & 27.9  & 8.9  & \textbf{11.3} & 12.0  & \textbf{26.7} & 11.6  & \textbf{14.5} & \textbf{7.9} & \textbf{22.9} & \textbf{57.4} \\
    \midrule
    \ \ No Prompts & 9.1  & 15.1$^{*}$ & 27.5  & 9.0  & 11.4  & 11.7$^{**}$ & 26.5  & 11.7  & 14.6  & 7.6$^{**}$ & 22.7  & 57.3  \\
    \midrule
    \midrule
       & \multicolumn{12}{c}{\textit{\textbf{Few-shot (50)}}} \\
    \midrule
    w/o GS & 26.5${_\text{0.2}}$ & 5.8${_\text{0.2}^{**}}$ & 17.7${_\text{0.5}^{**}}$ & 3.1${_\text{0.2}^{**}}$ & 26.3${_\text{0.5}}$ & 7.1${_\text{0.7}^{**}}$ & 20.1${_\text{1.0}^{**}}$ & 5.0${_\text{0.7}^{**}}$ & 21.1${_\text{0.1}}$ & 5.4${_\text{0.5}^{**}}$ & 16.5${_\text{0.8}^{**}}$ & 12.8${_\text{1.8}^{**}}$ \\
    Random & 101.6${_\text{25.1}}$ & 5.6${_\text{1.4}^{**}}$ & 16.9${_\text{2.6}^{**}}$ & 3.4${_\text{0.9}^{**}}$ & 51.3${_\text{16.4}}$ & 7.3${_\text{1.1}^{**}}$ & 19.9${_\text{2.6}^{**}}$ & 6.1${_\text{1.6}^{**}}$ & 72.0${_\text{18.8}}$ & 5.1${_\text{0.6}^{**}}$ & 16.0${_\text{0.8}^{**}}$ & 12.5${_\text{1.1}^{**}}$ \\
    Vocab & 13.5${_\text{0.2}}$ & 10.3${_\text{0.4}^{**}}$ & 20.9${_\text{0.2}}$ & 7.0${_\text{0.4}^{**}}$ & 20.6${_\text{0.9}}$ & 8.6${_\text{0.7}^{**}}$ & 21.5${_\text{0.7}}$ & 8.8${_\text{2.6}^{**}}$ & 20.8${_\text{0.2}}$ & 6.1${_\text{0.8}^{**}}$ & 17.4${_\text{0.6}^{**}}$ & 16.6${_\text{1.1}^{**}}$ \\
    Frequent & 13.9${_\text{0.1}}$ & 10.9${_\text{0.5}^{*}}$ & 21.0${_\text{0.2}}$ & 7.8${_\text{0.6}}$ & 21.2${_\text{0.9}}$ & 8.4${_\text{0.6}^{**}}$ & 21.5${_\text{0.6}}$ & 8.3${_\text{2.2}^{**}}$ & 21.3${_\text{0.2}}$ & 5.8${_\text{0.8}^{**}}$ & 17.3${_\text{0.7}^{**}}$ & 17.0${_\text{1.7}^{**}}$ \\
    Semantic & \textbf{13.3$_\text{0.1}$} & \textbf{11.1$_\text{0.4}$} & \textbf{21.3$_\text{0.3}$} & \textbf{8.3$_\text{0.4}$} & \textbf{20.1$_\text{0.5}$} & \textbf{9.0$_\text{0.6}$} & \textbf{22.0$_\text{0.6}$} & \textbf{10.1$_\text{2.6}$} & \textbf{20.0$_\text{0.2}$} & \textbf{6.4$_\text{0.7}$} & \textbf{18.7$_\text{0.5}$} & \textbf{29.0$_\text{2.6}$} \\
    \midrule
    \ \ w/o Co-Ind & 13.5${_\text{0.1}}$ & 10.4${_\text{0.2}^{**}}$ & 20.9${_\text{0.3}}$ & 6.8${_\text{0.2}^{**}}$ & 20.6${_\text{0.9}}$ & 8.8${_\text{0.7}}$ & 21.6${_\text{0.8}}$ & 8.7${_\text{2.6}^{**}}$ & -  & -  & -  & - \\
    \ \ w/o Sp-Ind & 14.2${_\text{0.1}}$ & 9.7${_\text{0.2}^{**}}$ & 19.9${_\text{0.2}^{**}}$ & 7.0${_\text{0.9}^{**}}$ & 21.1${_\text{0.7}}$ & 7.3${_\text{0.6}^{**}}$ & 20.8${_\text{0.8}^{**}}$ & 8.7${_\text{2.2}^{**}}$ & -  & -  & -  & - \\
    \ \ No Prompts & 14.4${_\text{0.2}}$ & 9.1${_\text{0.3}^{**}}$ & 19.6${_\text{0.2}^{**}}$ & 5.8${_\text{0.4}^{**}}$ & 21.4${_\text{0.9}}$ & 7.3${_\text{0.5}^{**}}$ & 20.8${_\text{0.7}^{**}}$ & 8.0${_\text{2.2}^{**}}$ & 21.1${_\text{0.1}}$ & 5.9${_\text{0.5}^{**}}$ & 18.1${_\text{0.3}}$ & 28.1${_\text{2.4}}$ \\
    \bottomrule
    \end{tabular}%
    }
  \caption{
  Results of continuous prompts.
  Subscripts denote standard deviations of eight different results.
  The \textbf{best results} under either the full-data or few-shot setting are \textbf{in bold}.
  $^{*}$/$^{**}$ denotes significant gaps to the best results ($p$-value $< 0.05$/$0.01$ respectively).
  These marks have the same meaning hereinafter.
  }
  \label{tab:continuous}%
  \vspace{-2mm}
\end{table*}%

\subsection{Evaluation Metrics}
\label{subsec:metric}

We adopted the following automatic metrics to evaluate the quality of model-generated responses.
\textbf{Perplexity (PPL)} \cite{zhang-etal-2018-personalizing} reflects the task adaptation ability by calculating the loss on the test samples.
\textbf{BLEU-$\bm{n}$} \cite{papineni-etal-2002-bleu, liu-etal-2021-towards} reflects the grammaticality and contextual coherence by computing the $n$-gram overlaps with golden responses.
We reported the corpus-level BLEU-2 (B-2) scores.
\textbf{Unigram F1} \cite{dinan2018wizard, Zhao2020Low-Resource} measures the lexical similarity between generated and golden responses (with NLTK tokenization).

To evaluate the groundedness of generation, we further used another two metrics.
For WoW and PC, we computed \textbf{Wiki/PSN F1} \cite{dinan2018wizard, shuster-etal-2021-retrieval-augmentation} as the unigram F1 between model-generated responses and the grounding sources (i.e., Wikipedia knowledge and persona).
Note that to truly reflect the informative referred contents, we only counted the non-stop words as overlapped unigrams.
For ESConv, we computed \textbf{Match Ratio} as the ratio of cases where the strategies identified from the generated responses exactly matched the designated ones\footnote{
ESConv defines 7 support strategies along with an ``others'' one, which does not explicitly refer to any specific strategy.
We removed the cases where the designated strategies are ``others'' when computing match ratio.
}.
To identify the responses' strategies, we fine-tuned a BERT-Large \cite{devlin-etal-2019-bert} classifier on the full training set of ESConv, which obtained 57.5 accuracy, 86.4 Hits@3 and 51.4 macro-F1 on the test set (8-class).

We conducted significance tests using bootstrap resampling \cite{berg-kirkpatrick-etal-2012-empirical} for BLEU, Students' t-test for F1 and Wiki/PSN F1, and sign test for Match Ratio.
Since the sample size affects statistical significance, for the few-shot experiments, we evenly sampled from the eight generation sets to construct a non-repetitive sample set for significance tests.

\section{Prompting GPT2}
\label{sec:gpt2}

In the implementation of our prompting method, \textit{continuous prompts} work via the newly added indicative tokens, while \textit{discrete prompts} introduce no new parameters but only textual descriptions.
While intuitively reasonable, the two types of prompts may have their own shortcomings under the few-shot learning setting, as revealed in previous works \cite{liu2021pre}.
Specifically, the initialization of continuous prompts could sensitively affect few-shot learning performance \cite{li-liang-2021-prefix, gu2021ppt}, and even minor perturbations on discrete prompts could lead to significant performance fluctuation \cite{schick-schutze-2021-shot}.
To address these concerns, in this section we take GPT2-medium as the backbone model, and in detail evaluate and analyze our method with different options (e.g., prompts' initialization or perturbation).

\begin{table*}[t]
  \centering
  \scalebox{0.75}{
    \begin{tabular}{l|cccc|cccc|cccc}
    \toprule
       & \multicolumn{4}{c|}{\textit{\textbf{Wizard-of-Wikipedia}}} & \multicolumn{4}{c|}{\textit{\textbf{PersonaChat}}} & \multicolumn{4}{c}{\textit{\textbf{ESConv}}} \\
\cmidrule{2-13}       & \textbf{PPL ↓} & \textbf{B-2} & \textbf{F1} & \textbf{Wiki F1} & \textbf{PPL ↓} & \textbf{B-2} & \textbf{F1} & \textbf{PSN F1} & \textbf{PPL ↓} & \textbf{B-2} & \textbf{F1} & \textbf{Match} \\
    \midrule
    \midrule
       & \multicolumn{4}{c|}{\textit{\textbf{Full-data (66K)}}} & \multicolumn{4}{c|}{\textit{\textbf{Full-data (50K)}}} & \multicolumn{4}{c}{\textit{\textbf{Full-data (12K)}}} \\
    \midrule
    Continuous & 9.0  & 15.3$^{**}$ & 27.9  & 8.9$^{**}$ & 11.3  & 12.0  & 26.7  & \textbf{11.6} & 14.5  & 7.9$^{**}$ & 22.9  & 57.4$^{**}$ \\
    Discrete & \textbf{9.0} & \textbf{15.9} & \textbf{28.1} & \textbf{9.4} & \textbf{11.2} & \textbf{12.2} & \textbf{26.9} & 11.4  & \textbf{14.4} & \textbf{8.2} & \textbf{23.4} & \textbf{59.9} \\
    \midrule
    \midrule
       & \multicolumn{12}{c}{\textit{\textbf{Few-shot (50)}}} \\
    \midrule
    Continuous & 13.3${_\text{0.1}}$ & 11.1${_\text{0.4}^{**}}$ & 21.3${_\text{0.3}^{**}}$ & 8.3${_\text{0.4}^{**}}$ & 20.1${_\text{0.5}}$ & \textbf{9.0$_\text{0.6}$} & 22.0${_\text{0.6}^{**}}$ & 10.1${_\text{2.6}^{**}}$ & 20.0${_\text{0.2}}$ & 6.4${_\text{0.7}^{**}}$ & 18.7${_\text{0.5}^{**}}$ & 29.0$_\text{2.6}^{**}$ \\
    Discrete & \textbf{12.1$_\text{0.1}$} & 11.9${_\text{0.5}}$ & 22.2${_\text{0.4}}$ & 8.9${_\text{0.5}^{**}}$ & 17.9${_\text{0.3}}$ & 8.9${_\text{0.7}}$ & 22.5${_\text{0.4}}$ & 10.3${_\text{1.6}}$ & \textbf{18.3$_\text{0.1}$} & 7.0${_\text{0.3}}$ & \textbf{20.0$_\text{0.4}$} & 42.2${_\text{3.6}^{**}}$ \\
    \midrule
    \ \ w/o Instruction & 12.1${_\text{0.1}}$ & 12.0${_\text{0.3}}$ & 22.2${_\text{0.3}}$ & 9.2${_\text{0.6}}$ & 17.9${_\text{0.2}}$ & 8.9${_\text{0.5}}$ & 22.6${_\text{0.5}}$ & 10.4${_\text{2.1}}$ & 18.4${_\text{0.1}}$ & \textbf{7.0$_\text{0.4}$} & 19.7${_\text{0.3}}$ & 43.1${_\text{4.3}^{**}}$ \\
    \ \ w/ Instruct-Pert & 12.2${_\text{0.1}}$ & \textbf{12.1$_\text{0.4}$} & \textbf{22.4$_\text{0.4}$} & \textbf{9.3$_\text{0.7}$} & 18.0${_\text{0.3}}$ & 8.6${_\text{0.5}^{**}}$ & 22.5${_\text{0.4}}$ & 10.6${_\text{1.9}}$ & 18.3${_\text{0.1}}$ & 6.9${_\text{0.4}}$ & 19.9${_\text{0.3}}$ & \textbf{44.5$_\text{4.3}$} \\
    \ \ w/ Speaker-Pert & 12.2${_\text{0.1}}$ & 12.1${_\text{0.3}}$ & 22.3${_\text{0.3}}$ & 9.1${_\text{0.5}}$ & \textbf{17.9$_\text{0.3}$} & 9.0${_\text{0.4}}$ & \textbf{22.8$_\text{0.4}$} & \textbf{10.7$_\text{1.9}$} & 18.4${_\text{0.1}}$ & 6.7${_\text{0.5}}$ & 19.8${_\text{0.3}}$ & 44.5${_\text{3.5}}$ \\
    \bottomrule
    \end{tabular}%
    }
  \caption{Results of discrete prompts.}
  \label{tab:discrete}%
  \vspace{-2mm}
\end{table*}%

\subsection{Continuous Prompts}
\label{subsec:continuous}

\noindent \textbf{Initialization Methods} \quad
Except random initialization\footnote{To ensure convergence, we trained the Random initialization method for 20 epochs (10 more than default).}, we compared three commonly used and intuitive ways of initializing the continuous prompts \cite{gu2021ppt}:
(1) using the pre-trained embedding of a random vocabulary token (Vocab), 
(2) using the pre-trained embedding of a random top-100 frequent token in the training corpus (Frequent),
and (3) using the average embeddings of the textual semantic explanations of the indicators (Semantic)\footnote{
For instance, for the token that indicates the components of knowledge or persona, we average the embeddings of the tokenized word ``knowledge'' or ''persona'' to initialize the corresponding indicative token.
}.
Note that on ESConv, the strategy tokens \cite{liu-etal-2021-towards} are initialized in the same way as continuous prompts.
We added a compared baseline as the control group where the GS is not provided (w/o GS).

\noindent \textbf{Results} \quad
Table \ref{tab:continuous} shows the results.
Unsurprisingly, on all three tasks, well initialized continuous prompts (e.g., Semantic) consistently boosts the performance compared to not using prompts (No Prompts).
Among different initialization methods, the Semantic initialization method performs consistently best and thus proves its soundness.
In contrast, random initialization (Random) shows dramatically poor performance, even worse than not adding GS (w/o GS), on all three tasks under the few-shot setting.

However, inserting continuous prompts or not or initializing them with different methods have only minor gaps on WoW and PC under the full-data setting.
Notably, semantic-initialized strategy tokens always bring much higher match ratios under both few-shot (compared to Random, 29.0 vs. 12.5) and full-data (57.4 vs. 40.6) settings, highlighting the necessity of leveraging the strategies' prior semantic meanings to achieve better controllability.

\noindent \textbf{Ablation Study} \quad
Based on the Semantic initialization, we ablated either the speaker (w/o Sp-Ind), the component (w/o Co-Ind) or both types of indicative tokens (No Prompts).
From Table \ref{tab:continuous}, they both contribute to the final prompting effects, showing the reasonableness of prompting GDG tasks by distinguishing the complex input constructs.
We notice that the speaker type occupies a larger contribution.
The reason may be that the speaker indicative tokens occur more in input sequences (up to 5 utterances) and thus provide major prompts about the constructs of input sequences.

\subsection{Discrete Prompts} 
\label{subsec:discrete}

\noindent \textbf{Prompt Perturbation} \quad
The basic discrete prompts are simply obtained by replacing indicators with textual explanations and prepending input sequences with task instructions (Figure \ref{fig:method}).
We also attempted three perturbations on the discrete prompts, as shown in Figure \ref{fig:discrete_pert}:
(1) removing task instructions (w/o Instruction),
(2) perturbing instructions by word substitution, including ``The following'' with ``Below'', ``conversation'' with ``dialog'', ``grounded'' with ``based'', etc. (w/ Instruct-Pert),
and (3) further substituting speaker words ``user''/``system'' with ``human''/``AI'' based on (2) (w/ Speaker-Pert).

\noindent \textbf{Results} \quad
Table \ref{tab:discrete} shows the results.
Profited by the prior information introduced in the textual descriptions, discrete prompts significantly outperform continuous prompts in most cases (except the B-2 metric on PC), and even maintain the remarkable superiority under the full-data setting (on WoW and ESConv).
Meanwhile, despite the need of manual crafting, the discrete prompts generally do not have obvious performance fluctuation due to minor modifications.
Such robustness to prompt perturbation comes from not only the all-parameter training (\S \ref{subsec:implementation}) \cite{logan2021cutting} but also the huge output space of GDG tasks (compared to NLU tasks).
It indicates the practicability of discrete prompts and that it may be not necessary to optimize them laboriously.



\begin{table*}[t]
  \centering
  \scalebox{0.75}{
    \begin{tabular}{l|cccc|cccc|cccc}
    \toprule
       & \multicolumn{4}{c|}{\textit{\textbf{Wizard-of-Wikipedia}}} & \multicolumn{4}{c|}{\textit{\textbf{PersonaChat}}} & \multicolumn{4}{c}{\textit{\textbf{ESConv}}} \\
\cmidrule{2-13}       & \textbf{PPL ↓} & \textbf{B-2} & \textbf{F1} & \textbf{Wiki F1} & \textbf{PPL ↓} & \textbf{B-2} & \textbf{F1} & \textbf{PSN F1} & \textbf{PPL ↓} & \textbf{B-2} & \textbf{F1} & \textbf{Match} \\
    \midrule
    \midrule
    \multicolumn{13}{c}{\textit{\textbf{GPT2-Medium (345M)}}} \\
    \midrule
    No Prompts & 14.4${_\text{0.2}}$ & 9.1${_\text{0.3}^{**}}$ & 19.6${_\text{0.2}^{**}}$ & 5.8${_\text{0.4}^{**}}$ & 21.4${_\text{0.9}}$ & 7.3${_\text{0.5}^{**}}$ & 20.8${_\text{0.7}^{**}}$ & 8.0${_\text{2.2}^{**}}$ & 21.1${_\text{0.1}}$ & 5.9${_\text{0.5}^{**}}$ & 18.1${_\text{0.3}^{**}}$ & 28.1${_\text{2.4}^{**}}$ \\
    Continuous & 13.3${_\text{0.1}}$ & 11.1${_\text{0.4}^{**}}$ & 21.3${_\text{0.3}^{*}}$ & 8.3${_\text{0.4}^{**}}$ & 20.1${_\text{0.5}}$ & \textbf{9.0$_\text{0.6}$} & 22.0${_\text{0.6}}$ & 10.1${_\text{2.6}}$ & 20.0${_\text{0.2}}$ & 6.4${_\text{0.7}^{**}}$ & 18.7${_\text{0.5}^{**}}$ & 29.0${_\text{2.6}^{**}}$ \\
    Discrete & \textbf{12.1$_\text{0.1}$} & \win{11.9$_\text{0.5}$} & \textbf{22.2$_\text{0.4}$} & \textbf{8.9$_\text{0.5}$} & \textbf{17.9$_\text{0.3}$} & 8.9${_\text{0.7}}$ & \textbf{22.5$_\text{0.4}$} & \textbf{10.3$_\text{1.6}$} & \textbf{18.3$_\text{0.1}$} & \win{7.0$_\text{0.3}$} & \win{20.0$_\text{0.4}$} & \win{42.2$_\text{3.6}$} \\
    \midrule
    \midrule
    \multicolumn{13}{c}{\textit{\textbf{T5-Base (220M)}}} \\
    \midrule
    No Prompts & \textbf{13.7$_\text{0.1}$} & 9.2${_\text{0.3}^{**}}$ & 20.3${_\text{0.3}}$ & \textbf{10.8$_\text{0.7}$} & \textbf{13.4$_\text{0.1}$} & \textbf{8.6$_\text{0.6}$} & \textbf{22.5$_\text{0.1}$} & \textbf{16.0$_\text{0.8}$} & \textbf{22.3$_\text{0.1}$} & 4.7${_\text{0.5}^{**}}$ & 16.1${_\text{0.4}^{**}}$ & 13.8${_\text{0.8}^{**}}$ \\
    Continuous & 13.8${_\text{0.1}}$ & 9.0${_\text{0.3}^{**}}$ & 20.3${_\text{0.2}}$ & 10.4${_\text{0.3}}$ & 13.6${_\text{0.1}}$ & 8.4${_\text{0.7}^{*}}$ & 22.4${_\text{0.3}}$ & 15.5${_\text{0.8}^{*}}$ & 22.5${_\text{0.2}}$ & 4.5${_\text{0.4}^{**}}$ & 15.8${_\text{0.3}^{**}}$ & 13.4${_\text{0.4}^{**}}$ \\
    Discrete & 14.6${_\text{0.3}}$ & \textbf{10.0$_\text{0.3}$} & \textbf{20.7$_\text{0.4}$} & 8.2${_\text{0.7}^{**}}$ & 14.4${_\text{0.3}}$ & 5.6${_\text{0.6}^{**}}$ & 18.8${_\text{0.9}^{**}}$ & 11.1${_\text{1.5}^{**}}$ & 23.7${_\text{0.2}}$ & \textbf{6.2$_\text{0.2}$} & \textbf{17.6$_\text{0.2}$} & \textbf{25.0$_\text{0.9}$} \\
    \midrule
    \midrule
    \multicolumn{13}{c}{\textit{\textbf{T5-Large (770M)}}} \\
    \midrule
    No Prompts & 11.1${_\text{0.1}}$ & 10.5${_\text{0.1}^{**}}$ & 22.2${_\text{0.2}^{**}}$ & 12.3${_\text{0.5}^{**}}$ & 10.5${_\text{0.2}}$ & 8.1${_\text{0.7}^{**}}$ & 23.6${_\text{0.3}^{**}}$ & 15.9${_\text{1.2}^{**}}$ & 17.8${_\text{0.1}}$ & 5.1${_\text{0.5}^{**}}$ & 17.2${_\text{0.4}^{**}}$ & 16.8${_\text{0.5}^{**}}$ \\
    Continuous & 10.7${_\text{0.1}}$ & \textbf{10.9$_\text{0.2}$} & 22.6${_\text{0.2}^{**}}$ & 11.5${_\text{0.8}^{**}}$ & 10.4${_\text{0.1}}$ & 10.1${_\text{0.8}}$ & 23.8${_\text{0.3}}$ & 18.8${_\text{1.8}}$ & 17.4${_\text{0.1}}$ & 5.4${_\text{0.5}^{**}}$ & 17.4${_\text{0.3}^{**}}$ & 16.4${_\text{0.8}^{**}}$ \\
    Discrete & \textbf{10.5$_\text{0.1}$} & 10.9${_\text{0.3}}$ & \win{23.8$_\text{0.2}$} & \win{13.6$_\text{0.9}$} & \textbf{10.1$_\text{0.1}$} & \win{10.4$_\text{1.1}$} & \win{24.6$_\text{0.2}$} & \win{18.9$_\text{2.5}$} & \textbf{16.9$_\text{0.1}$} & \textbf{6.4$_\text{0.6}$} & \textbf{19.4$_\text{0.4}$} & \textbf{38.9$_\text{2.0}$} \\
    \midrule
    \midrule
    \multicolumn{13}{c}{\textit{\textbf{BART-Large (400M)}}} \\
    \midrule
    No Prompts & \textbf{16.1$_\text{0.4}$} & 9.0${_\text{0.4}^{*}}$ & 20.5${_\text{0.6}}$ & \textbf{11.8$_\text{1.2}$} & 26.5${_\text{1.4}}$ & \textbf{8.4$_\text{1.5}$} & 21.1${_\text{0.8}}$ & 14.4${_\text{2.7}^{**}}$ & 23.3${_\text{0.5}}$ & \textbf{6.3$_\text{0.7}$} & 17.4${_\text{0.5}}$ & 19.4${_\text{0.9}}$ \\
    Continuous & 16.4${_\text{0.3}}$ & 8.9${_\text{0.8}}$ & 19.9${_\text{0.7}^{*}}$ & 8.9${_\text{1.4}^{**}}$ & \textbf{25.6$_\text{1.1}$} & 7.9${_\text{1.0}^{**}}$ & \textbf{21.2$_\text{0.8}$} & 13.1${_\text{1.9}^{**}}$ & \textbf{22.9$_\text{0.6}$} & 6.2${_\text{0.9}}$ & 17.8${_\text{0.7}}$ & \textbf{21.6$_\text{2.0}$} \\
    Discrete & 16.3${_\text{0.6}}$ & \textbf{9.6$_\text{0.6}$} & \textbf{20.6$_\text{0.7}$} & 10.4${_\text{2.2}^{**}}$ & 25.8${_\text{0.7}}$ & 8.1${_\text{1.5}}$ & 20.6${_\text{1.2}}$ & \textbf{15.7$_\text{3.8}$} & 25.0${_\text{0.4}}$ & 6.0${_\text{0.6}}$ & \textbf{18.0$_\text{0.8}$} & 15.9${_\text{1.8}^{**}}$ \\
    \midrule
    \midrule
    \multicolumn{13}{c}{\textit{\textbf{DialoGPT-Medium (345M)}}} \\
    \midrule
    No Prompts & \textbf{56.8${_\text{1.1}}$} & \textbf{6.7$_\text{0.4}$} & 18.7${_\text{0.8}}$ & 4.9${_\text{0.6}^{**}}$ & \textbf{17.7$_\text{0.2}$} & \textbf{9.6$_\text{0.5}$} & \textbf{23.2$_\text{0.3}$} & 10.9${_\text{1.2}^{**}}$ & 35.2${_\text{1.0}}$ & 5.3${_\text{0.4}^{**}}$ & 18.6${_\text{0.4}^{*}}$ & \textbf{30.6$_\text{1.2}$} \\
    Continuous & 60.1${_\text{2.4}}$ & 6.5${_\text{0.6}}$ & 18.6${_\text{0.8}}$ & 5.1${_\text{0.8}}$ & 19.2${_\text{0.4}}$ & 9.4${_\text{0.5}}$ & 22.6${_\text{0.4}}$ & 10.7${_\text{1.4}^{**}}$ & \textbf{34.7${_\text{1.4}}$} & 5.4${_\text{0.3}^{**}}$ & 18.6${_\text{0.3}^{*}}$ & 27.1${_\text{3.1}^{**}}$ \\
    Discrete & 95.5${_\text{2.7}}$ & 6.4${_\text{0.4}}$ & 16.8${_\text{0.5}^{**}}$ & \textbf{5.3$_\text{0.8}$} & 24.1${_\text{0.8}}$ & 9.3${_\text{0.7}}$ & 23.1${_\text{0.5}}$ & \textbf{11.8$_\text{2.0}$} & 42.8${_\text{1.4}}$ & \textbf{5.8$_\text{1.0}$} & \textbf{19.1$_\text{0.4}$} & 26.7${_\text{3.9}^{**}}$ \\
    \midrule
    \midrule
    \multicolumn{13}{c}{\textit{\textbf{Blender-Small (90M)}}} \\
    \midrule
    No Prompts & \textbf{14.8$_\text{0.2}$} & \textbf{9.8$_\text{0.3}$} & \textbf{21.6$_\text{0.4}$} & 8.6${_\text{1.3}}$ & 15.0${_\text{0.2}}$ & 8.5${_\text{1.0}^{**}}$ & 22.9${_\text{0.6}}$ & 9.5${_\text{1.3}^{**}}$ & 21.0${_\text{0.2}}$ & 6.1${_\text{0.7}}$ & 19.5${_\text{0.6}}$ & \textbf{34.9$_\text{2.8}$} \\
    Continuous & 14.9${_\text{0.1}}$ & 9.6${_\text{0.4}}$ & 21.3${_\text{0.5}}$ & 8.2${_\text{0.8}^{*}}$ & \textbf{14.9$_\text{0.2}$} & 8.5${_\text{0.8}^{**}}$ & \textbf{23.0$_\text{0.2}$} & 9.7${_\text{0.8}}$ & 21.1${_\text{0.2}}$ & \textbf{6.1$_\text{0.6}$} & \textbf{19.6$_\text{0.4}$} & 34.7${_\text{3.1}}$ \\
    Discrete & 15.4${_\text{0.3}}$ & 9.0${_\text{0.9}^{**}}$ & 20.8${_\text{0.9}^{*}}$ & \textbf{8.7$_\text{0.7}$} & 14.9${_\text{0.3}}$ & \textbf{9.4$_\text{0.7}$} & 22.9${_\text{0.6}}$ & \textbf{10.0$_\text{1.5}$} & \textbf{20.1${_\text{0.1}}$} & 5.7${_\text{0.5}^{**}}$ & 18.6${_\text{0.6}^{**}}$ & 15.7${_\text{1.9}^{**}}$ \\
    \bottomrule
    \end{tabular}%
    }
  \caption{
  Results of using different pre-trained models.
  The best results among all the models are \win{highlighted}.
  Perplexity (PPL) is not comparable between models due to the differences in vocabulary and tokenization.
  }
  \label{tab:comparison}%
  \vspace{-1mm}
\end{table*}%

\section{Prompting Different Pre-trained Models}
\label{sec:comparison}

While both continuous and discrete prompts perform well with GPT2, we further wonder whether they still work with other backbone models.
As shown in previous works \cite{liu2021gpt, li-liang-2021-prefix}, effective prompting methods vary from the adopted pre-trained models, depending on the pre-training settings, model architectures, etc.
In this section, we thoroughly investigate how our prompting method works with different pre-trained models, and analyze the factors that influence the effectiveness of prompting.

\subsection{Compared Models}

We compared several representative pre-trained models to be used as the backbone models, which are also popularly adopted in previous works of dialog generation \cite{mi2021cins, shuster-etal-2021-retrieval-augmentation}.
Note that our choice of model sizes was mainly limited by computational resources (Tesla V100 32G), and we used the largest available and feasible models within the range that resources allow.

\noindent \textbf{Language Models} \quad
\textbf{GPT2-Medium} (345M parameters) \cite{radford2019language} is an \textit{autoregressive} language model, pre-trained with the \textit{language modeling} objective.
We also included three \textit{encoder-decoder} language models, \textbf{T5-Base} (220M), \textbf{T5-Large} (770M\footnote{
While the further enlarged GPT2 (Large, 762M) has a similar parameter number to T5-Large, the architecture of GPT2 leads to much more GPU memory occupation, overloading our computational resources.
Thus the largest GPT2 we could experiment with is GPT2-Medium.
}) \cite{raffel2020exploring} and \textbf{BART-Large} (400M) \cite{lewis-etal-2020-bart}.
T5 and BART both adopt the \textit{denoising} objectives but are different in terms of the noising functions, the input/output formats and the training corpora.

\noindent \textbf{Conversational Models} \quad
We meanwhile included \textbf{DialoGPT-Medium} (345M) \cite{zhang-etal-2020-dialogpt} and \textbf{Blender-Small} (90M) \cite{roller-etal-2021-recipes}.
They are both pre-trained on massive Reddit corpora while Blender is further fine-tuned on several crowd-sourced datasets\footnote{
Since we found that the fine-tuning of Blender does not utilize the grounding sources \cite{roller-etal-2021-recipes}, we still experimented with Blender for comparison.
} \cite{dinan2018wizard, zhang-etal-2018-personalizing, rashkin-etal-2019-towards, smith-etal-2020-put}.
Note that we did not include larger-sized Blender models because they only accept the encoder input of max length 128.

We compared three methods with different pre-trained models: not using prompts (No Prompts), continuous prompts and discrete prompts.
The input formats of DialoGPT are the same as GPT2, and those of T5 and BART/Blender are shown in Figure \ref{fig:method_t5} and \ref{fig:method_bart} respectively. 
Results are shown in Table \ref{tab:comparison}.

\subsection{Language Models vs. Conversational Models}
\label{subsec:vs}

Prompted language models perform superiorly to conversational models on all three tasks, while unprompted ones generally perform worse than the latter.
On WoW and ESConv, prompted GPT2 achieves higher B-2, F1 and Match Ratio scores than both DialoGPT and Blender, but unprompted GPT2, T5 and BART all underperform Blender.
On PC, prompting also enables T5-Large to outperform DialoGPT in terms of all the metrics.
We think that such observation is not trivial and rather important.
It suggests that only pre-training on massive general dialog corpora makes it difficult for conversational models to make use of non-dialog external information.
In contrast, although not specially pre-trained on dialog corpora, language models (e.g., GPT2 and T5) can still be quickly adapted to dialog generation tasks, and at the same time acquire the ability to utilize external information, that is, the capability of grounding.

While critical to language models (except BART, as will be discussed later), prompting does not work with conversational models.
A direct evidence is the PPLs of DialoGPT (Discrete $>$ Continuous $\approx$ No Prompts, on all three tasks).
Intuitively, \textit{discrete prompts} are naturally not tailored for conversational models due to the enormous gaps with the input formats of pre-training (i.e., concatenating utterances of the conversation context as the encoder input).
Hence, discrete prompts would instead hurt conversational models' performance and are usually inferior to continuous prompts or not using prompts on all three tasks.
As for \textit{continuous prompts}, they seem to differ little from the performance of not adding prompts.
We conjecture that the reason is that conversational models have been able to distinguish the input constructs during pre-training on dialog corpora, where the conversation context could contain utterances from multiple speakers.


\subsection{Further Analysis}
\label{subsec:analysis}

\noindent \textbf{Effects of Model Pre-training} \quad
Among the three language models, GPT2 and T5-Large both benefit from continuous and discrete prompts, while BART generally does not (only small improvements on WoW).
It probably results from the differences in their pre-training objectives and corpora.
Specifically, GPT2 and T5 adopt the more general pre-training objectives (language modeling and span corruption, respectively) than BART (denoising in an autoencoding way), and T5 is even pre-trained on much larger pre-training corpora (745GB vs. BART's 160GB).
As a result, GPT2 and T5 can be more easily stimulated by well initialized continuous prompts and natural language discrete prompts.

\noindent \textbf{Effects of Model Sizes} \quad
Comparing T5 of two sizes, we notice that unlike T5-Large, T5-Base usually is not profited by continuous prompts on all three tasks, and its performance is even damaged by discrete prompts on PC.
It suggests that the larger model size is beneficial to effective prompting, which is also our motivation to experiment with as large as possible pre-trained models.

\noindent \textbf{Effects of Model Architectures} \quad
Continuous and discrete prompts benefit both GPT2 and T5-Large.
It indicates that our prompting method is effective with language models of different architectures.

Interestingly, we note that encoder-decoder models are more prone to copy GS than autoregressive language models, that is, T5 and BART achieve notably higher Wiki/PSN F1 than GPT2 on WoW and PC.
We hypothesize that this phenomenon results from the different model architectures.
Specifically, given that the GS is positioned before the conversation context, the bidirectional encoding of T5 and BART enables the unified attention to the model input. 
In contrast, the unidirectional attention of GPT2 may focus more on the contents nearby to the target responses (i.e., the conversation context rather than the GS).

\begin{table}[t]
  \centering
  \scalebox{0.75}{
    \begin{tabular}{l|cc|cc}
    \toprule
       & \multicolumn{2}{c|}{\textit{\textbf{Wizard-of-Wikipedia}}} & \multicolumn{2}{c}{\textit{\textbf{PersonaChat}}} \\
\cmidrule{2-5}       & \textbf{B-2} & \textbf{Wiki F1} & \textbf{B-2} & \textbf{PSN F1} \\
    \midrule
    \midrule
    \multicolumn{5}{c}{\textit{\textbf{GPT2-Medium (345M)}}} \\
    \midrule
    No Prompts & 9.1 {\small (\lose{+0.0})} & 9.0 {\small (\win{+3.3})} & 7.5 {\small (+0.2)} & 12.6 {\small (+4.6)} \\
    Continuous & 10.7 {\small (-0.4)} & 11.1 {\small (+2.8)} & 9.2 {\small (+0.2)} & 15.5 {\small (\win{+5.5})} \\
    Discrete & 9.4 {\small (\win{-2.6})} & 6.8 {\small (-2.1)} & 7.8 {\small (-0.8)} & 10.6 {\small (+0.3)} \\
    \midrule
    \midrule
    \multicolumn{5}{c}{\textit{\textbf{T5-Base (220M)}}} \\
    \midrule
    No Prompts & 8.6 {\small (-0.6)} & 9.3 {\small (-1.4)} & 8.2 {\small (-0.5)} & 15.3 {\small (-0.6)} \\
    Continuous & 7.7 {\small (-1.3)} & 10.2 {\small (-0.2)} & 8.4 {\small (\lose{+0.0})} & 14.9 {\small (-0.5)} \\
    Discrete & 8.2 {\small (-1.8)} & 6.6 {\small (-1.7)} & 5.2 {\small (-0.5)} & 10.2 {\small (-0.9)} \\
    \midrule
    \midrule
    \multicolumn{5}{c}{\textit{\textbf{T5-Large (770M)}}} \\
    \midrule
    No Prompts & 9.8 {\small (-0.7)} & 11.5 {\small (-0.8)} & 7.3 {\small (-0.7)} & 14.8 {\small (-1.1)} \\
    Continuous & 10.0 {\small (-0.9)} & 11.5 {\small (\lose{-0.0})} & 9.2 {\small (-0.9)} & 17.2 {\small (-1.6)} \\
    Discrete & 9.6 {\small (-1.2)} & 11.4 {\small (-2.2)} & 9.2 {\small (\win{-1.2})} & 18.0 {\small (-0.9)} \\
    \midrule
    \midrule
    \multicolumn{5}{c}{\textit{\textbf{BART-Large (400M)}}} \\
    \midrule
    No Prompts & 8.9 {\small (-0.1)} & 10.0 {\small (-1.9)} & 8.2 {\small (-0.2)} & 13.6 {\small (-0.8)} \\
    Continuous & 9.0 {\small (+0.1)} & 9.2 {\small (+0.2)} & 7.4 {\small (-0.6)} & 12.9 {\small (\lose{-0.2})} \\
    Discrete & 9.3 {\small (-0.2)} & 11.0 {\small (+0.6)} & 7.9 {\small (-0.2)} & 15.4 {\small (-0.3)} \\
    \bottomrule
    \end{tabular}%
    }
  \caption{
  Results of post-positioning the GS right after the conversation context.
  Performance changes are in parentheses.
  The largest and smallest \textbf{changes} (absolute values) among all the models are \win{highlighted} and \lose{shadowed} respectively.
  }
  \label{tab:postposition}%
  \vspace{-2mm}
\end{table}%

To verify our hypothesis, comparing the same model by modifying the attention directions seems direct but instead infeasible, because it would perturb the pre-trained models, especially under the few-shot setting.
Alternatively, we moved the GS right after the conversation context (the corresponding discrete prompts are shown in Figure \ref{fig:discrete_post}), aiming to observe whether generated responses refer more to the GS.
Results are shown in Table \ref{tab:postposition}.
For the No Prompts and Continuous methods, post-GS makes GPT2 achieve largely increased Wiki/PSN F1, while T5 and BART are not influenced obviously.
It implies the effects of model architectures and indirectly proves the reasonableness of our hypothesis.
Meanwhile, we note that the performance of discrete prompts drops remarkably with both GPT2 and T5, indicating that post-positioning the GS is not a suitable prompt design for these two language models.

\section{Conclusion}
This work explores the prompt-based few-shot learning for grounded dialog generation (GDG).
We show that distinguishing the constructs of model input is effective to boost the few-shot learning performance, in which well initialized continuous prompts or easily designed discrete prompts play the key role.
We additionally demonstrate that our prompting method performs well with language models of different architectures (e.g., GPT2 and T5) but does not work with conversational models (e.g., DialoGPT and Blender), among which prompted language models can even achieve superior performance to conversational models.
Further analysis shows that the effectiveness of our prompting method also relies on backbone models with enough prowess.
Our work reveals the potential of prompting methods in the few-shot learning for GDG, and raises attention to the proper selection of pre-trained models in GDG tasks.

\section*{Acknowledgements}

This work was supported by the National Science Foundation for Distinguished Young Scholars (with No. 62125604) and the NSFC projects (Key project with No. 61936010 and regular project with No. 61876096). This work was also supported by the Guoqiang Institute of Tsinghua University, with Grant No. 2019GQG1 and 2020GQG0005.

\bibliography{anthology,custom}
\bibliographystyle{acl_natbib}

\appendix

\begin{figure}[t]
  \centering
  \includegraphics[width=0.9\linewidth]{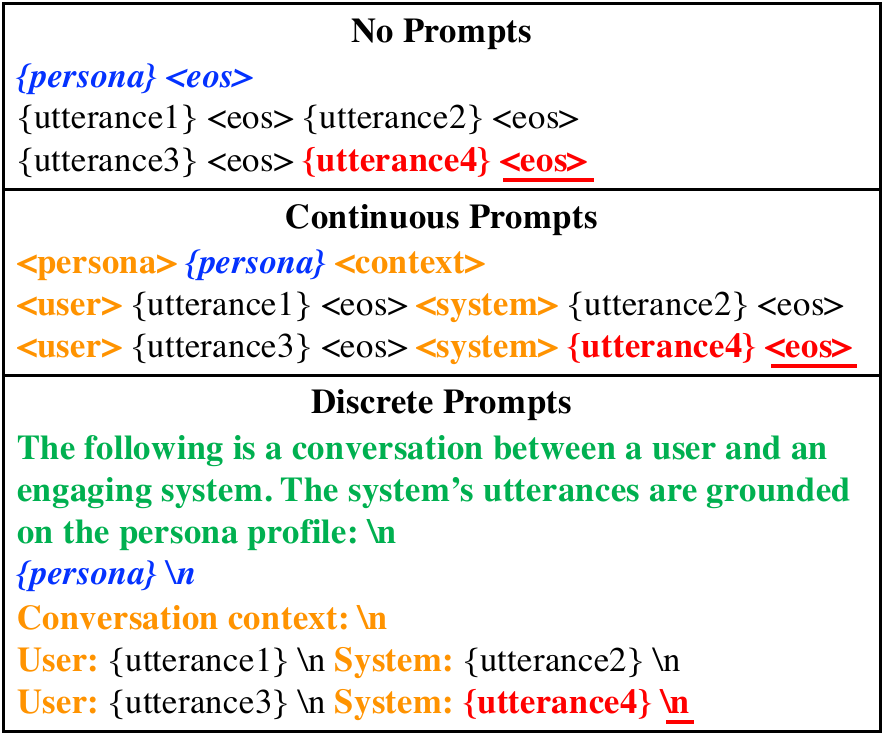}
  \caption{
  Applying our prompting method to PersonaChat.
  }
  \label{fig:method_pc}
\end{figure}

\begin{figure}[t]
  \centering
  \includegraphics[width=0.9\linewidth]{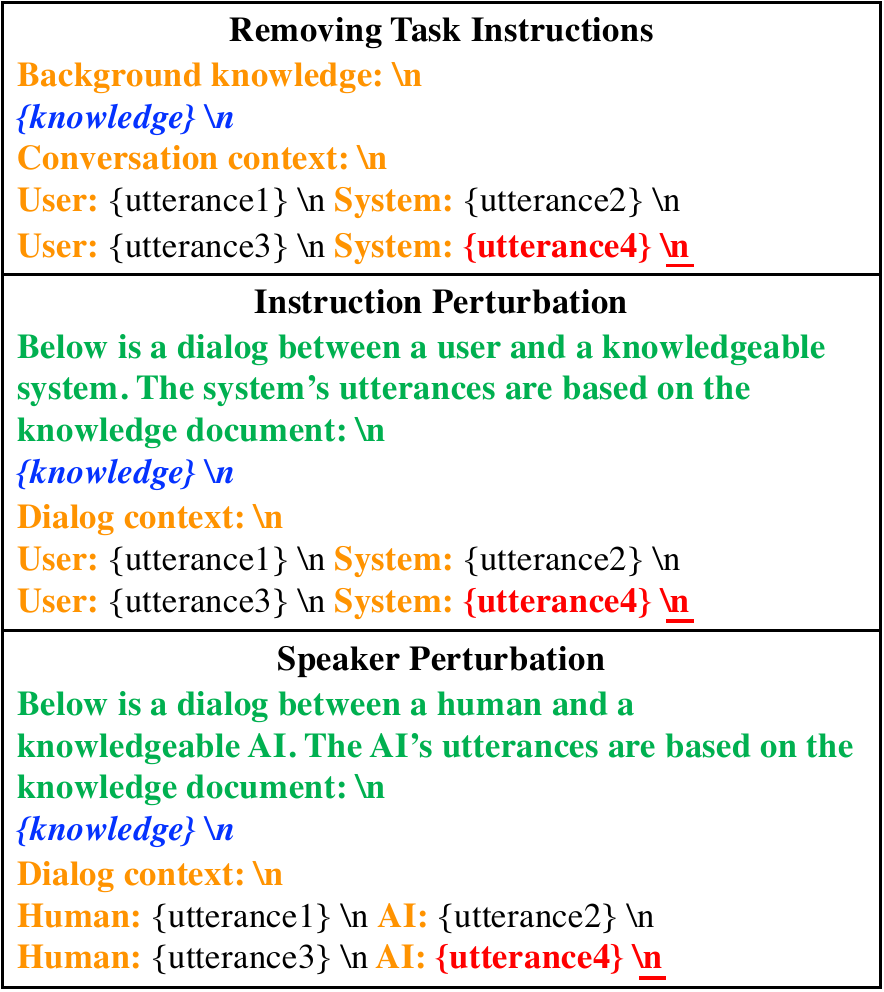}
  \caption{
  Perturbed discrete prompts.
  }
  \label{fig:discrete_pert}
\end{figure}

\begin{figure*}[t]
  \centering
  \includegraphics[width=0.9\linewidth]{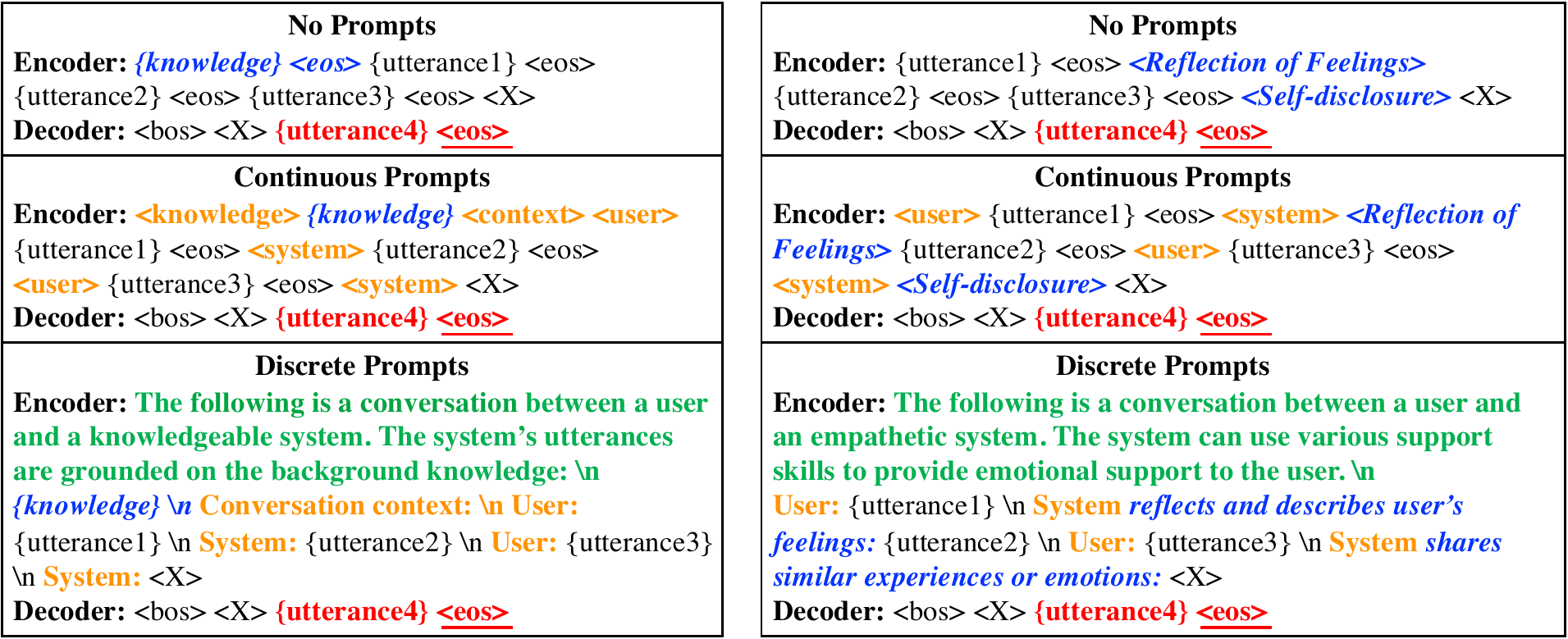}
  \caption{
  Input formats for T5.
  }
  \label{fig:method_t5}
\end{figure*}

\begin{figure*}[t]
  \centering
  \includegraphics[width=0.9\linewidth]{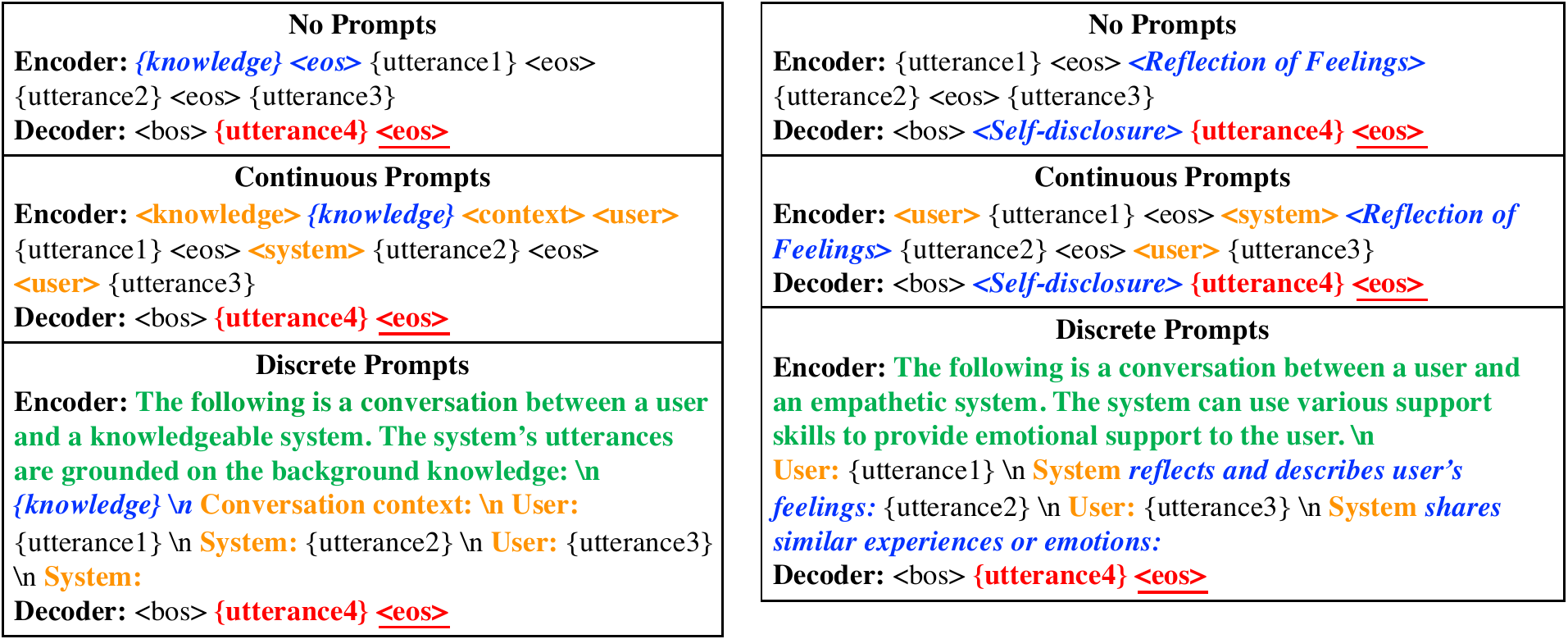}
  \caption{
  Input formats for BART and Blender.
  }
  \label{fig:method_bart}
\end{figure*}

\begin{figure}[t]
  \centering
  \includegraphics[width=0.9\linewidth]{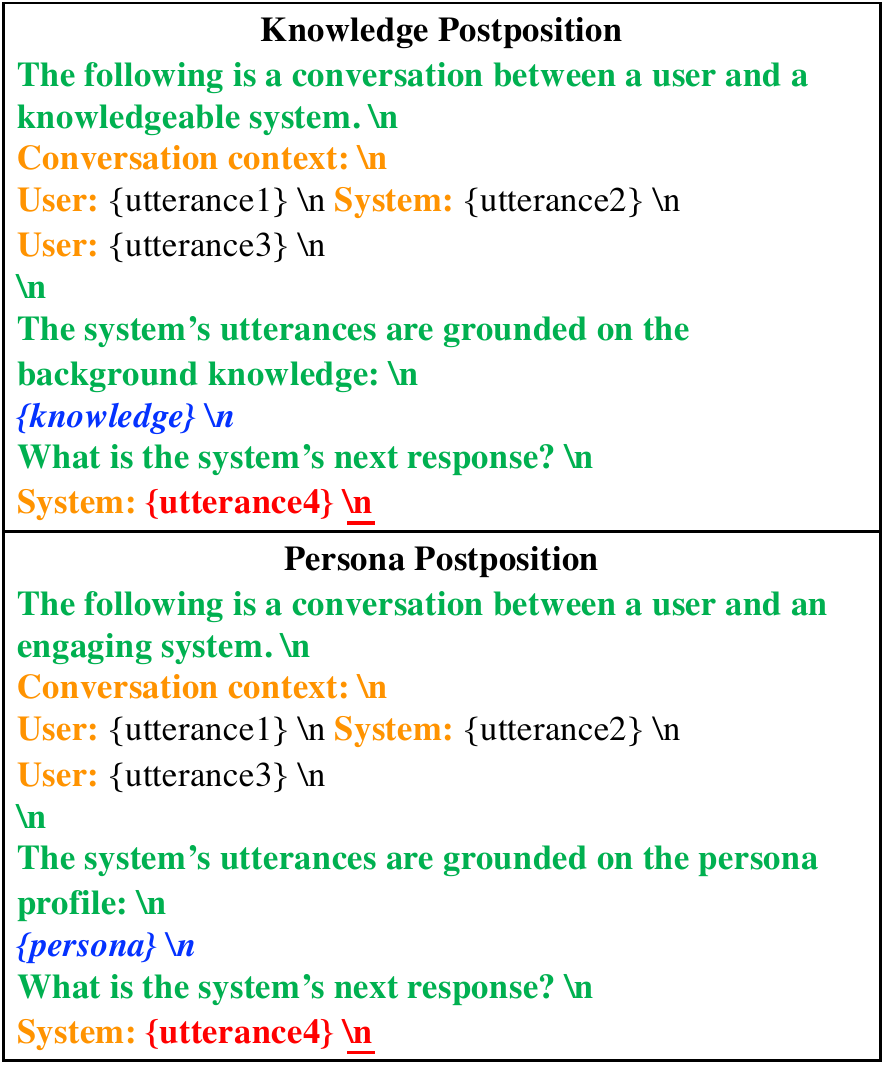}
  \caption{
  Post-positioning the GS in discrete prompts.
  }
  \label{fig:discrete_post}
\end{figure}

\section{Used Pre-trained Models}

This work experiments with the following open-sourced pre-trained models: BERT-Large\footnote{\url{https://huggingface.co/bert-large-uncased}}, GPT2-Medium\footnote{\url{https://huggingface.co/gpt2-medium}}, T5-Base\footnote{\url{https://huggingface.co/t5-base}}, T5-Large\footnote{\url{https://huggingface.co/t5-large}}, BART-Large\footnote{\url{https://huggingface.co/facebook/bart-large}}, DialoGPT-Medium\footnote{\url{https://huggingface.co/microsoft/DialoGPT-medium}} and Blender-Small\footnote{\url{https://huggingface.co/facebook/blenderbot_small-90M}}.

\end{document}